\documentclass[11pt]{article}

\usepackage[utf8]{inputenc}
\usepackage{lmodern}  
\usepackage[T1]{fontenc}
\usepackage{amsmath,amssymb,amsfonts}
\usepackage{graphicx}
\usepackage{booktabs}
\usepackage{hyperref}
\usepackage{url}
\usepackage{xcolor}
\usepackage{listings}
\usepackage{microtype}
\usepackage[margin=1in]{geometry}
\usepackage{caption}
\usepackage{subcaption}
\usepackage{float}

\hypersetup{
  colorlinks=true,
  linkcolor=blue!60!black,
  citecolor=blue!60!black,
  urlcolor=blue!60!black,
}

\title{Remember, Don't Re-read: Stateful ReAct Agents for Token-Efficient Autonomous Experimentation}

\author{
  Faramarz Jabbarvaziri \\
  \texttt{faramarz.vaziri87@gmail.com}
}

\date{}

\begin{document}
\maketitle

\begin{abstract}
The autoresearch pattern enables autonomous experimentation by having a large language model (LLM) iteratively modify code to optimize a target metric.
Its stateless design, however, reconstructs experimental context from scratch at every iteration, incurring $O(n)$ token cost per iteration and $O(n^{2})$ total.
This work reformulates the pattern as a stateful ReAct agent using LangGraph, where typed persistent state carries experimental history across iterations via a tool-calling interface.
Two benchmarks are evaluated: hyperparameter tuning (15 iterations, small per-iteration observations) and code performance optimization (40 iterations, large per-iteration observations containing full source code and benchmark results).
On hyperparameter tuning, the stateful agent consumes 90\% fewer tokens (2{,}492 vs.\ 24{,}465).
On code optimization, the stateful agent consumes 52\% fewer tokens (627K vs.\ 1{,}275K) while achieving comparable optimization quality on both tasks.
The token reduction is structural: the stateless agent re-reads the full history at $O(n)$ cost per iteration, while the stateful agent operates within a fixed-size conversation window at $O(1)$ cost.
This paper describes the architecture in sufficient detail for practitioners to implement a stateful autoresearch agent for their own workflows.
\end{abstract}

\section{Introduction}

Machine learning research is fundamentally iterative: a researcher proposes a hypothesis, implements a code change, runs an experiment, analyzes results, and decides what to try next.
This cycle repeats dozens to hundreds of times before convergence.
While the computational cost of individual experiments has decreased, the human reasoning between experiments remains the primary bottleneck.

Karpathy~\cite{karpathy2026} showed that an LLM can close this loop autonomously.
The \emph{autoresearch} pattern decomposes experimentation into three files: a frozen data pipeline (\texttt{prepare.py}), a mutable training script (\texttt{train.py}), and natural-language research directives (\texttt{program.md}).
An LLM reads the directives, modifies the training script, executes the experiment, observes metrics, and repeats.
Karpathy in its initial demonstration, this system ran approximately 700 experiments over two days and discovered 20 independent training optimizations without human intervention.

The elegance of this pattern lies in its simplicity: the LLM's only interface with the ML system is code modification, and the only feedback signal is the target metric.
However, this simplicity introduces a structural inefficiency.
Each LLM invocation is \emph{stateless}---the agent must reconstruct experimental context by re-reading the full results history accumulated at program.md file at every iteration.
As the number of experiments grows, the prompt grows linearly, consuming tokens to re-transmit information the agent has already processed.
This is not a prompt engineering problem; Huang et al.~\cite{huang2025latent} show that LLMs cannot internally maintain transient state across interactions, making external state management an architectural necessity.

In this work, the autoresearch pattern is reformulated as a \textbf{stateful ReAct agent}~\cite{yao2023react} using LangGraph.
The agent interacts with the ML system through tool calls rather than monolithic prompts, and experimental history, strategic reasoning, and convergence tracking persist in a typed state graph across iterations.
The key structural advantage is that per-iteration token cost drops from $O(n)$ to $O(1)$, enabling arbitrarily long experiment sequences without prompt truncation or summarization.

\section{Related Work}

\subsection{LLM Agents for ML Research}

The autoresearch pattern belongs to a growing family of agentic AI systems that prioritize ease of implementation and flexibility over provably optimal search.
Unlike classical AutoML, which operates within predefined parameterized search spaces, LLM-guided experimentation treats the search space itself as mutable---the agent can modify arbitrary code rather than selecting from a fixed set of configurations.

Several systems explore this space.
MLAgentBench~\cite{huang2024mlagentbench} provides a 13-task benchmark for evaluating ML agents; the AI Scientist~\cite{lu2024aiscientist,lu2025aiscientistv2} extends scope to full paper generation; AIDE~\cite{jiang2025aide} frames ML engineering as tree search over code variants; AgentHPO~\cite{liu2024agenthpo} applies LLM agents to hyperparameter optimization, matching human performance on 12 tasks; Agent Laboratory~\cite{schmidgall2025agentlab} demonstrates a multi-agent pipeline at 84\% lower cost.
The autoresearch pattern~\cite{karpathy2026} takes the most minimal approach---a single LLM iteratively editing a training script---but its stateless design limits efficiency on longer experiment sequences.
The present work preserves this minimality while adding persistent state.

\subsection{Reasoning and Memory in LLM Agents}

ReAct~\cite{yao2023react} interleaves reasoning traces with tool actions.
Reflexion~\cite{shinn2023reflexion} extends this with explicit self-reflection and episodic memory.
Huang et al.~\cite{huang2025latent} show that LLMs cannot maintain transient state across interactions, motivating external memory.
MemGPT~\cite{packer2023memgpt} addresses this with OS-inspired memory hierarchies; Voyager~\cite{wang2023voyager} builds a skill library for embodied agents; CoALA~\cite{sumers2024coala} provides a taxonomy of agent memory types.
The present work instantiates these principles for iterative experimentation: the graph state serves as working memory, the experiment history as episodic memory, and the domain constraints as semantic memory, managed through typed state transitions rather than LLM-driven paging.

\section{Method}

\subsection{Architecture}

\begin{figure}[H]
  \centering
  \includegraphics[width=0.4\textwidth]{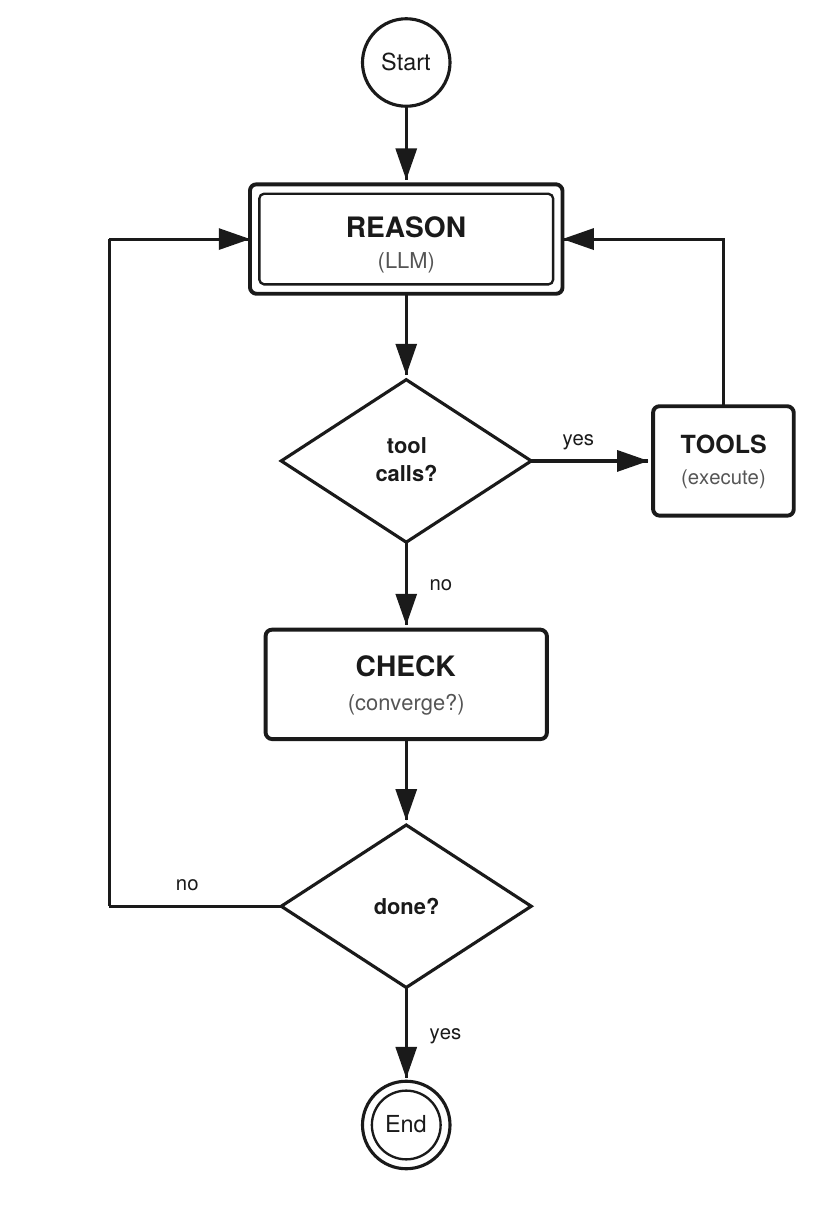}
  \caption{ReAct agent state graph. \textsc{Reason} invokes the LLM; if it produces tool calls, control passes to \textsc{Tools} and returns. Otherwise, \textsc{Check} evaluates convergence. If not converged, control loops back to \textsc{Reason}.}
  \label{fig:state-graph}
\end{figure}

The agent is implemented as a LangGraph state graph with three node types (Figure~\ref{fig:state-graph}).

\textbf{REASON} invokes the LLM (Claude Haiku 4.5) with the current state context.
The model generates reasoning traces and tool calls following the ReAct pattern~\cite{yao2023react}.
\textbf{TOOLS} executes tool calls deterministically and returns observations.
\textbf{CHECK} evaluates convergence criteria without LLM involvement.
Conditional edges route based on whether the LLM produced tool calls (to \textsc{Tools}) or a terminal response (to \textsc{Check}).

The graph is constructed using \texttt{langgraph.graph.StateGraph} with conditional edges:

\begin{lstlisting}[language=Python]
graph = StateGraph(AgentState)
graph.add_node("reason", reason_node)
graph.add_node("tools", ToolNode(TOOLS))
graph.add_node("check", check_node)

graph.set_entry_point("reason")

# After reason: tool calls -> tools, else -> check
graph.add_conditional_edges(
    "reason",
    lambda s: "tools" if has_tool_calls(s)
              else "check",
)
# After tools -> back to reason
graph.add_edge("tools", "reason")
# After check -> continue or end
graph.add_conditional_edges(
    "check",
    lambda s: "end" if s["status"] != "running"
              else "reason",
)
\end{lstlisting}

\subsection{State Schema}

The agent state is defined as a typed dictionary with six fields:

\begin{lstlisting}[language=Python]
class AgentState(TypedDict):
    messages: list
    experiment_history: list
    current_best: dict
    iteration: int
    train_py_content: str
    status: str
\end{lstlisting}

This state persists across all iterations.
Unlike the stateless design, where the LLM must reconstruct context from the results table at each step, the stateful agent carries forward its experimental history, current best result, and reasoning trace.
The \texttt{messages} field is bounded by a sliding window of 20 messages, ensuring that per-iteration input cost does not grow with experiment count.

An \texttt{ExperimentState} helper class manages the history and provides a compact summary to the LLM at each step:

\begin{lstlisting}[language=Python]
class ExperimentState:
    def __init__(self):
        self.history: list[dict] = []
        self.best_f1: float = 0.0
        self.best_params: dict = {}
        self.best_iteration: int = -1
        self.iteration: int = 0
        self.tried_configs: set = set()
        self.strategy_notes: list[str] = []

    def summary(self) -> str:
        """Compact state summary for the LLM."""
        lines = [
          f"Iteration: {self.iteration}/{MAX}",
          f"Best F1: {self.best_f1}",
        ]
        # Only last 5 experiments (state carries
        # the rest implicitly)
        recent = self.history[-5:]
        for h in recent:
            lines.append(
              f"  Iter {h['iteration']}: "
              f"F1={h['f1']}, "
              f"params={json.dumps(h['params'])}"
            )
        return "\n".join(lines)
\end{lstlisting}

The \texttt{summary()} method is what makes the $O(1)$ cost possible: regardless of how many experiments have been run, the LLM receives only the most recent 5, plus aggregate statistics.
The full history remains accessible in state but does not enter the prompt.

\subsection{Tools and Guardrails}

The agent accesses the ML system through two tools: \texttt{get\_experiment\_history} (query past results) and \texttt{run\_experiment} (train with specified hyperparameters, return metrics).
Following the autoresearch principle that the data pipeline must be immutable, the agent can modify only hyperparameters, not data loading or evaluation code.

In the production variant (see Section~\ref{sec:production-agent}), additional tools are available: \texttt{get\_current\_train\_py} to read the current training notebook, \texttt{modify\_train\_py} to upload a modified version with guardrails (must preserve evaluation and logging calls), \texttt{submit\_training\_job} to trigger a Databricks job and poll until completion, and \texttt{query\_results} to run read-only SQL against the results table.

\subsection{Convergence}

The agent terminates when any of three conditions holds: (1) the target metric crosses a user-defined threshold, (2) the iteration budget is exhausted, or (3) the LLM concludes that further improvement is unlikely.

\section{Experiments}

Two benchmark tasks are used to evaluate the stateful architecture across different observation sizes per iteration.
Both use Claude Haiku 4.5 via the Databricks model serving endpoint (\texttt{databricks-claude-haiku-4-5}), 3 random seeds (42, 123, 456), and the same stateless-vs-stateful comparison structure.
Token counts are estimated via tiktoken (\texttt{cl100k\_base}).
All experiments ran on Databricks serverless compute.

\subsection{Task 1: Hyperparameter Tuning}
\label{sec:task-hparam}

The first task optimizes XGBoost hyperparameters on the UCI Covertype dataset~\cite{blackard1999covertype} from the University of California, Irvine Machine Learning Repository---a 7-class forest cover classification task with 54 features, subsampled to 20{,}000 instances.
Each iteration produces a small observation: a JSON hyperparameter configuration (${\sim}$200 tokens) and three scalar metrics (F1, precision, recall).
The budget is 15 iterations.

This task represents the \textbf{small-observation regime}: the stateless agent's per-iteration prompt grows by ${\sim}$200 tokens per experiment, so the total cost grows as $\sum_{i=1}^{n} 200i = O(n^{2})$ but with a small constant.

\paragraph{Stateless runner.}
At each iteration, the runner constructs a fresh prompt containing the \emph{complete} experiment history---every iteration's parameters and metrics---plus the current best score.
This prompt is sent as a single message to the LLM; the response is parsed for a JSON hyperparameter configuration.

\paragraph{Stateful runner.}
The stateful runner maintains an \texttt{ExperimentState} object and a persistent conversation across all iterations.
At each step, the LLM receives only a compact summary (current iteration, best score, last 5 experiments, strategy notes) rather than the full history.
The message window is trimmed to the most recent 20 messages.

\begin{table}[H]
\centering
\small
\caption{Task 1: Hyperparameter tuning results (15 iterations, mean over 3 seeds). The F1 difference is within noise; the token reduction is the primary finding.}
\label{tab:hparam-results}
\begin{tabular}{@{}lccc@{}}
\toprule
\textbf{Metric} & \textbf{Stateless} & \textbf{Stateful} & \textbf{Ratio} \\
\midrule
Best macro F1 (mean $\pm$ std) & 0.764 $\pm$ 0.006 & 0.779 $\pm$ 0.004 & --- \\
Total tokens (mean) & 24{,}465 & 2{,}492 & \textbf{9.8$\times$} \\
\quad Input tokens & 19{,}885 & 1{,}317 & 15.1$\times$ \\
\quad Output tokens & 4{,}580 & 1{,}175 & 3.9$\times$ \\
Wall-clock time (mean) & 116\,s & 373\,s & 0.3$\times$ \\
\bottomrule
\end{tabular}
\end{table}

\begin{figure}[H]
  \centering
  \includegraphics[width=0.85\textwidth]{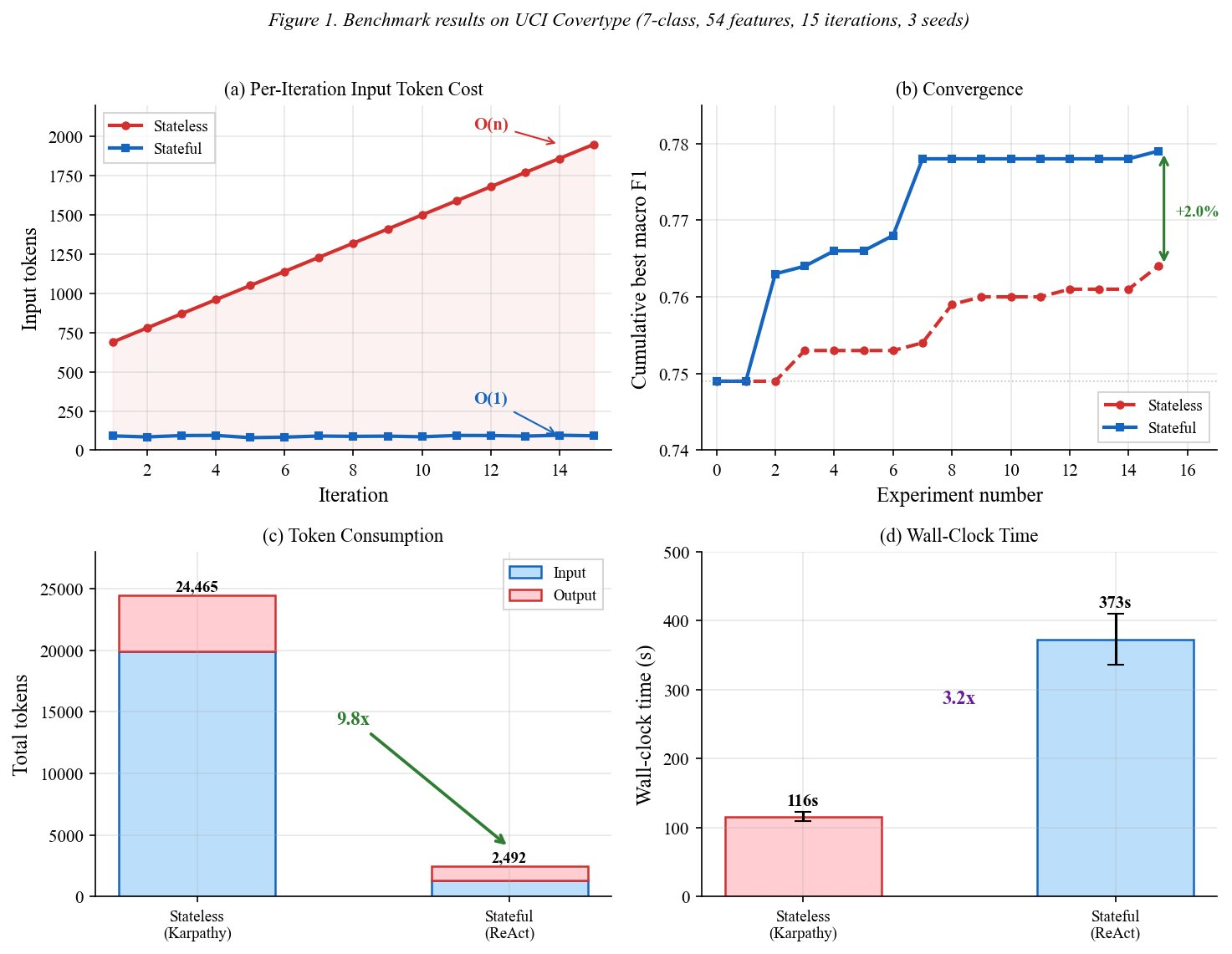}
  \caption{Task 1: Hyperparameter tuning on UCI Covertype (15 iterations, 3 seeds).
  (a)~Per-iteration input token cost: stateless grows linearly; stateful remains constant at $O(1)$.
  (b)~Cumulative best macro F1 over iterations.
  (c)~Total token consumption: 9.8$\times$ reduction.
  (d)~Wall-clock time comparison.}
  \label{fig:hparam-benchmark}
\end{figure}

In the small-observation regime, the stateful agent achieves a \textbf{9.8$\times$ token reduction} (Table~\ref{tab:hparam-results}).
Both agents reach comparable F1 scores; the observed difference (0.779 vs.\ 0.764) is within seed variance and is not a claimed contribution.
The per-iteration input token cost for the stateless agent grows linearly from ${\sim}$700 at iteration 1 to ${\sim}$1{,}900 at iteration 15 (Figure~\ref{fig:hparam-benchmark}a), while the stateful agent remains flat at ${\sim}$100 tokens per iteration.

\subsection{Task 2: Code Performance Optimization}
\label{sec:task-codeopt}

The second task optimizes a deliberately inefficient Python data-processing function (100 lines, processing 10{,}000 employee records) for execution speed.
The function contains bubble sorts, string concatenation in loops, manual dictionary accumulation, and other anti-patterns that an LLM can systematically replace with Pythonic constructs (\texttt{collections.Counter}, \texttt{sorted()}, list comprehensions, f-strings).
Each iteration produces a \emph{large} observation: the complete modified source code (${\sim}$1{,}500--3{,}000 tokens) plus benchmark results (median execution time, speedup, correctness check).
The budget is 40 iterations.

This task represents the \textbf{large-observation regime}: the stateless agent's per-iteration prompt grows by ${\sim}$2{,}000--4{,}000 tokens per experiment.
Over 40 iterations, the cumulative history sent at iteration 40 exceeds 80{,}000 tokens---approaching context window limits for smaller models.

\paragraph{Stateless runner.}
At each iteration, the runner constructs a fresh prompt containing the \emph{complete} optimization history: every previous code version, its benchmark timing, speedup factor, and correctness status.
The prompt follows the structure:

\begin{lstlisting}
Full optimization history:

=== Iteration 0 (time=8.14ms, speedup=1.00x) ===
Code:
def process_records(records):
    filtered = []
    for r in records:
        if r["active"] == True and r["years"] >= 3:
            ...
Benchmark:
Execution time: 8.140 ms (median of 20 runs)
Baseline time:  8.140 ms
Speedup:        1.00x

=== Iteration 1 (time=4.55ms, speedup=1.79x) ===
Code:
def process_records(records):
    filtered = [r for r in records
                if r["active"] and r["years"] >= 3]
    ...
Benchmark:
Execution time: 4.550 ms
Speedup:        1.79x
...
(all 39 previous iterations)

Propose the next optimized version.
\end{lstlisting}

By iteration 40, this history contains ${\sim}$40 full code listings and exceeds 80{,}000 tokens of input per call.
The response is parsed for a \texttt{CODE:} block containing the modified function, which is executed in a sandboxed \texttt{exec()} call.
The function must return the correct output structure (verified against required keys) before its timing is recorded.

\paragraph{Stateful runner.}
The stateful runner maintains a \texttt{CodeState} object and a persistent conversation (\texttt{messages} list) across all iterations.
The \texttt{CodeState} tracks:

\begin{lstlisting}[language=Python]
class CodeState:
    def __init__(self):
        self.history = []
        self.best_time_ms = float("inf")
        self.best_code = ""
        self.best_iteration = -1
        self.iteration = 0
        self.strategy_notes = []
        self.failed_approaches = []
\end{lstlisting}

At each step, the LLM receives a compact summary plus only the current best code and the latest benchmark result---not the full history:

\begin{lstlisting}
Iteration: 25/40
Baseline: 8.140 ms
Best: 4.275 ms (1.90x, iter 6)
Recent:
  Iter 21: 4.516ms (1.80x) [OK]
  Iter 22: 4.290ms (1.90x) [OK] *BEST*
  Iter 23: 4.713ms (1.73x) [OK]
  Iter 24: 4.633ms (1.76x) [OK]
  Iter 25: 4.688ms (1.74x) [OK]
Strategies: use Counter for tags; replace
  bubble sort with sorted(); join for strings
Failed: iter 7 (missing keys in output)

Current best code:
def process_records(records):
    ...

Propose an optimized version.
\end{lstlisting}

The LLM responds with both a \texttt{STRATEGY:} line (a one-sentence hypothesis about the proposed optimization) and a \texttt{CODE:} block.
The strategy note is stored in \texttt{state.strategy\_notes}; if the code fails correctness checks, the approach is recorded in \texttt{state.failed\_approaches} and the agent reverts to the best known code.
After benchmarking, the result is fed back as a short message (e.g., ``Result: 4.290ms (1.90x). NEW BEST!''), and the conversation continues.
The message window is trimmed to the most recent 20 messages to prevent context overflow.

\begin{table}[H]
\centering
\small
\caption{Task 2: Code optimization results (40 iterations, mean over 3 seeds). Both conditions achieve comparable speedup; the stateful agent uses half the tokens.}
\label{tab:codeopt-results}
\begin{tabular}{@{}lccc@{}}
\toprule
\textbf{Metric} & \textbf{Stateless} & \textbf{Stateful} & \textbf{Ratio} \\
\midrule
Best speedup (mean $\pm$ std) & 1.87$\times$ $\pm$ 0.02 & 1.91$\times$ $\pm$ 0.03 & --- \\
Total tokens (mean) & 1{,}275{,}309 & 626{,}702 & \textbf{2.0$\times$} \\
\quad Input tokens & 1{,}245{,}477 & 600{,}161 & 2.1$\times$ \\
\quad Output tokens & 29{,}832 & 26{,}541 & 1.1$\times$ \\
Failed iterations (of 40) & 0.7 & 1.3 & --- \\
\bottomrule
\end{tabular}
\end{table}

\begin{figure}[H]
  \centering
  \includegraphics[width=0.85\textwidth]{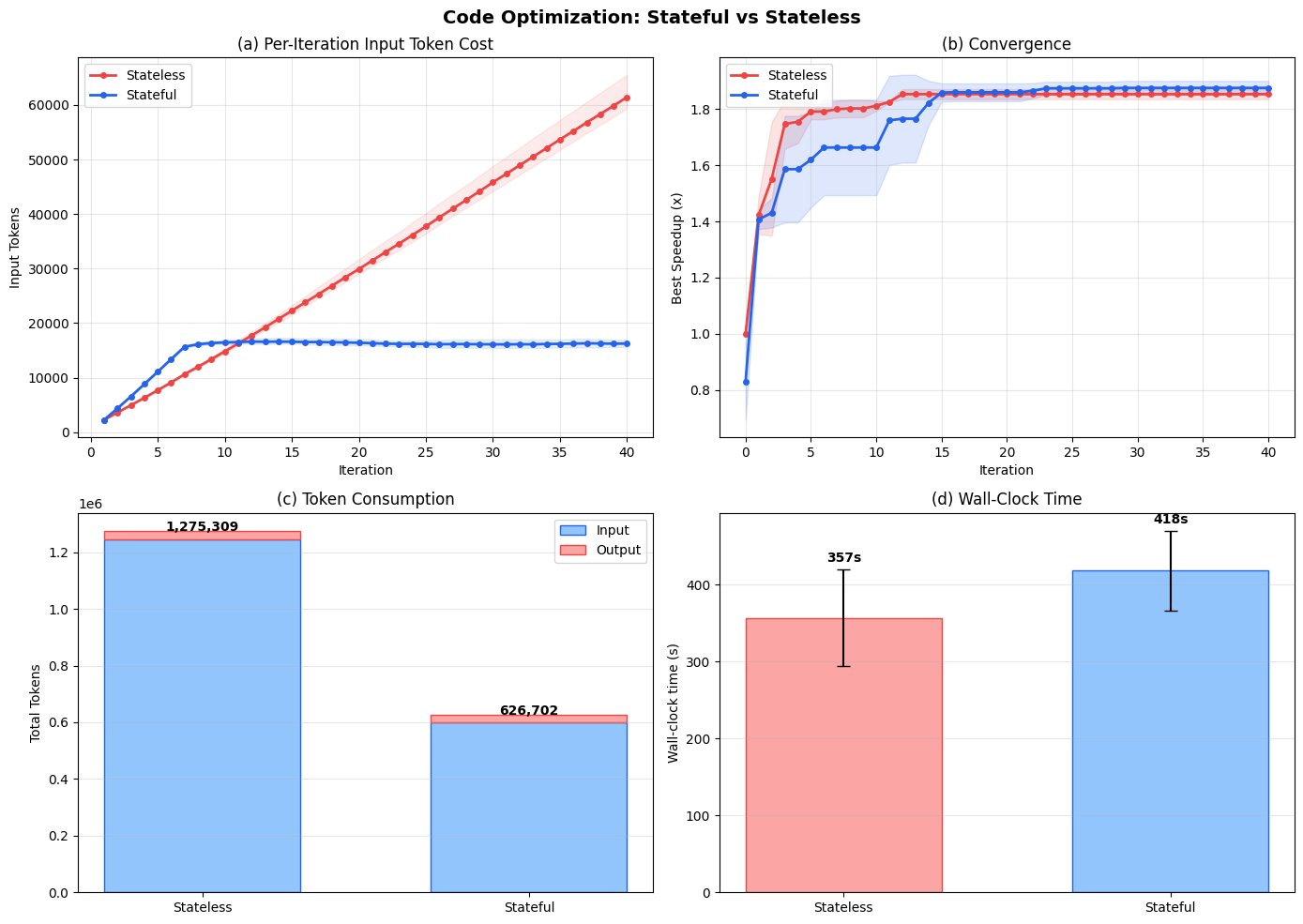}
  \caption{Task 2: Code performance optimization (40 iterations, 3 seeds).
  (a)~Per-iteration input token cost: the stateless prompt grows linearly as code listings accumulate, while the stateful agent plateaus once the 20-message window saturates (carrying only the current best code and recent results).
  (b)~Best speedup over the unoptimized baseline; both conditions converge to ${\sim}$1.9$\times$.
  (c)~Total token consumption: 2.0$\times$ reduction (1{,}275{,}309 vs.\ 626{,}702).
  (d)~Wall-clock time comparison.}
  \label{fig:codeopt-benchmark}
\end{figure}

In the large-observation regime, the stateful agent achieves a \textbf{2.0$\times$ token reduction} (Table~\ref{tab:codeopt-results}; Figure~\ref{fig:codeopt-benchmark}).
The ratio is lower than the 9.8$\times$ seen in Task~1 because the stateful agent must still send the current best code (${\sim}$1{,}500 tokens) at every iteration---a fixed cost that does not shrink with state management.
As shown in Figure~\ref{fig:codeopt-benchmark}a, the stateful agent's per-iteration input cost rises during early iterations and then plateaus once the message window saturates, whereas the stateless cost continues to climb linearly through iteration 40.
Both agents achieve comparable optimization quality, reaching ${\sim}$1.9$\times$ speedup over the deliberately unoptimized baseline by replacing bubble sorts with \texttt{sorted()}, string concatenation with \texttt{str.join()}, and manual loops with \texttt{collections.defaultdict} and list comprehensions.

\subsection{Comparing the Two Regimes}

\begin{table}[H]
\centering
\small
\caption{Token reduction ratio as a function of observation size per iteration. The stateful advantage is largest when per-iteration observations are small relative to the state summary.}
\label{tab:regime-comparison}
\begin{tabular}{@{}lcccc@{}}
\toprule
\textbf{Task} & \textbf{Iterations} & \textbf{Observation size} & \textbf{Stateless tokens} & \textbf{Reduction} \\
\midrule
Hyperparameter tuning & 15 & ${\sim}$200 tok & 24{,}465 & \textbf{9.8$\times$} \\
Code optimization & 40 & ${\sim}$3{,}000 tok & 1{,}275{,}309 & \textbf{2.0$\times$} \\
\bottomrule
\end{tabular}
\end{table}

The two tasks illustrate how the token savings depend on the ratio of observation size to state summary size (Table~\ref{tab:regime-comparison}).
When observations are small (a few hundred tokens of hyperparameters and metrics), the stateful agent's fixed-size summary is dramatically smaller than the accumulated history, yielding a 9.8$\times$ reduction in just 15 iterations.
When observations are large (full source code at ${\sim}$3{,}000 tokens each), the stateful agent still sends the current code every iteration, so the fixed cost is higher and the ratio is 2.0$\times$ at 40 iterations.
In both cases, the stateless cost grows as $O(n^{2})$ total while the stateful cost grows as $O(n)$, so the ratio continues to widen with additional iterations.

\section{Production Agent Architecture}
\label{sec:production-agent}

The benchmark runners demonstrate the core stateless-vs-stateful distinction, but the full production agent (used for internal ML research) extends the pattern with richer tooling and integration with Databricks.
This section describes the production architecture as a reference for practitioners.

\subsection{Tools}

The production agent exposes five tools to the LLM via LangGraph's \texttt{ToolNode}:

\begin{enumerate}
  \item \texttt{get\_experiment\_history()}: Queries a Delta table for the 20 most recent experiment results, returning metrics, parameters, and timestamps via the Databricks SQL Statement API.
  \item \texttt{get\_current\_train\_py()}: Exports the current training notebook from the Databricks workspace as source code, so the agent can inspect what it is about to modify.
  \item \texttt{modify\_train\_py(new\_content, experiment\_name, description)}: Uploads a modified training notebook with guardrails---the tool rejects submissions that remove the \texttt{evaluate\_model()}, \texttt{log\_result()}, or \texttt{save\_model\_to\_mlflow()} calls, preventing the agent from breaking the evaluation pipeline.
  \item \texttt{submit\_training\_job()}: Triggers a Databricks job via the Jobs API and polls until completion (60-second intervals, 1-hour timeout). On success, fetches and returns the latest result row from the results table.
  \item \texttt{query\_results(sql)}: Executes read-only SQL against Databricks tables, allowing the agent to compare experiments, inspect data distributions, or check for overfitting between validation and test splits.
\end{enumerate}

\subsection{System Prompt}

The system prompt defines the agent's domain context: the target metric, what can and cannot be changed in the training code, and the rules of engagement (one experiment per iteration, unique experiment names, mandatory evaluation calls).
This prompt is fixed across all iterations and does not grow with history.

\subsection{Graph Nodes}

The \texttt{reason\_node} binds the tools to the LLM via \texttt{ChatAnthropic.bind\_tools(TOOLS)} and invokes the model with the full system prompt plus the accumulated message history.
The \texttt{check\_node} increments the iteration counter, updates \texttt{current\_best} if the latest experiment improved the target metric, and sets \texttt{status} to \texttt{"converged"} or \texttt{"max\_iterations"} as appropriate.
The \texttt{end\_node} appends a summary message to the conversation.

\section{Analysis}

\subsection{Scaling Properties}

The token reduction is not merely a cost optimization.
At Haiku pricing (\$0.80/MTok input, \$4/MTok output), the per-run cost difference on Task~1 is modest.
However, Task~2 demonstrates that the gap becomes material with larger observations and longer sequences: 1.25M tokens for stateless versus 627K for stateful represents a meaningful cost difference at scale---hundreds of experiments per day across multiple projects.

More importantly, the $O(n^{2})$ total cost of stateless execution versus $O(n)$ for the stateful agent means the ratio continues to widen with experiment count.
Extrapolating from Task~2: at 100 iterations with 3{,}000-token observations, the stateless agent would consume ${\sim}$6M input tokens versus ${\sim}$1.5M for stateful---a 4$\times$ ratio.
At 200 iterations, the ratio would exceed 7$\times$.
The $O(1)$ per-iteration property also means the stateful agent can run arbitrarily long experiment sequences without prompt truncation or summarization, preserving decision-making capacity regardless of history length.

\begin{table*}[H]
\centering
\caption{Comparison of architectural properties between the stateless autoresearch baseline and the stateful LangGraph agent.}
\label{tab:comparison}
\begin{tabular}{@{}lp{5.5cm}p{5.5cm}@{}}
\toprule
\textbf{Dimension} & \textbf{Stateless}~\cite{karpathy2026} & \textbf{Stateful} \\
\midrule
Token scaling & $O(n)$ per iter; $O(n^2)$ total & $O(1)$ per iter; $O(n)$ total \\
Memory model & Ephemeral & Persistent typed state \\
Interface & Monolithic prompt $\to$ JSON & Tool-calling: reason $\to$ act $\to$ observe \\
Convergence & External (iteration budget or scripted threshold) & Integrated (metric + budget + agent judgment) \\
Audit trail & Results table only & Full state checkpoint \\
\bottomrule
\end{tabular}
\end{table*}

\subsection{Limitations}

The evaluation has several limitations.
First, two tasks are evaluated (hyperparameter tuning and code optimization); a more comprehensive study would include additional domains such as prompt optimization, scientific formula discovery, and hardware design generation, where the autoresearch pattern has also been applied~\cite{karpathy2026}.
Second, the wall-clock overhead of the ReAct pattern (3.2$\times$ on Task~1) may be prohibitive for tasks with very fast training loops, though the multi-turn conversation overhead is amortized over longer experiments.
Third, a single LLM (Haiku) is used; larger models with stronger in-context learning may partially compensate for the lack of persistent state.
Finally, 3 seeds provide limited statistical power; the optimization quality differences between conditions are within noise and should not be interpreted as quality advantages.

\section{Conclusion}

This work shows that reformulating the autoresearch pattern~\cite{karpathy2026} as a stateful ReAct agent yields a structural efficiency gain: $O(1)$ per-iteration token cost versus $O(n)$.
On two benchmark tasks spanning different observation sizes, the stateful agent reduces total token consumption by 9.8$\times$ (small observations, 15 iterations) and 2.0$\times$ (large observations, 40 iterations) while achieving comparable optimization quality.
The advantage grows with experiment sequence length and is independent of the underlying task.
Future work should extend to longer experiment horizons and additional domains where the quadratic token cost of stateless execution becomes the dominant bottleneck.


\end{document}